# Spatial-Spectral Hyperspectral Classification based on Learnable 3D Group Convolution

Guandong Li[1*], Mengxia Ye[2]

1.*Suning, Xuanwu, Nanjing, 210042, Jiangsu, China.

2.Aegon THTF, Qinghuai, Naijing, 210042, Jiangsu, China

*Corresponding author(s). E-mail(s): leeguandon@gmail.com

*Abstract:* Deep neural networks have faced many problems in hyperspectral image classification, including the ineffective utilization of spectral-spatial joint information and the problems of gradient vanishing and overfitting that arise with increasing depth. In order to accelerate the deployment of models on edge devices with strict latency requirements and limited computing power, this paper proposes a learnable group convolution network (LGCNet) based on an improved 3D-DenseNet model and a lightweight model design. The LGCNet module improves the shortcomings of group convolution by introducing a dynamic learning method for the input channels and convolution kernel grouping, enabling flexible grouping structures and generating better representation ability. Through the overall loss and gradient of the backpropagation network, the 3D group convolution is dynamically determined and updated in an end-to-end manner. The learnable number of channels and corresponding grouping can capture different complementary visual features of input images, allowing the CNN to learn richer feature representations. When extracting high-dimensional and redundant hyperspectral data, the 3D convolution kernels also contain a large amount of redundant information. The LGC module allows the 3D-DenseNet to choose channel information with more semantic features, and is very efficient, making it suitable for embedding in any deep neural network for acceleration and efficiency improvements. LGC enables the 3D-CNN to achieve sufficient feature extraction while also meeting speed and computing requirements. Furthermore, LGCNet has achieved progress in inference speed and accuracy, and outperforms mainstream hyperspectral image classification methods on the Indian Pines, Pavia University, and KSC datasets.

Key words: HSI classification; neural networks;3d group convolution; spatial-spectral information; 3d-densenet; lightweight

1. Introduction

Hyperspectral images typically consist of dozens to hundreds of spectral bands, which are narrowband images that provide nearly continuous spectral information and enhance the discrimination ability of land features [1]. In addition to collecting spectral information, hyperspectral imagers also gather spatial geometric distribution information of each sample point, forming a three-dimensional data cube that integrates two-dimensional spatial information with spectral information, known as "image and spectral combined". Land classification is an essential component of various applications of hyperspectral remote sensing, and it has been widely used in environmental monitoring, traffic planning, crop yield estimation, land use management, and engineering surveying, among others [2].

In recent years, deep learning has become one of the most successful technologies and has achieved impressive performance in the field of computer vision. The introduction of deep learning into remote sensing field for classification of hyperspectral images (HSI) has been influenced by this kind of technology [3-4]. Compared with traditional artificial feature design methods, deep learning can automatically learn advanced abstract features from complex hyperspectral data. Liu et al. [5]

proposed an effective classification framework based on deep learning and active learning, using DBN to extract deep spectral features, and adopting active learning algorithm to iteratively select high-quality labeled samples as training samples. Lee et al. [6-7] first used PCA to dimensionally reduce the entire hyperspectral data, and then input the spatial information contained in the neighboring area of the hyperspectral data, using 2D CNN for feature extraction. The method combines PCA and CNN, not only extracting effective spatial features but also reducing computational cost. In addition, Zhao et al. [8] proposed a classification framework based on spectral-spatial feature classification (SSFC) to classify HSIs. In this framework, spectral and spatial features were extracted through BLDE and CNN respectively, and then spectral and spatial features were fused to train a multi-feature classifier. Yue et al. [9] used deep CNN to extract deep features and combined it with logistic regression classifier for classification. However, methods like 2D-CNN [10-11] need to separately extract spatial and spectral information to extract features, which cannot fully utilize the joint spatial and spectral information and require complex preprocessing. Zhong et al. [12] based on 3D-CNN, proposed to use supervised spectral residual network (SSRN), to design continuous spatial and spectral residual modules respectively to extract spatial and spectral information. Based on the double convolutional pooling structure [13], directly using stacked 3D-CNN for classification achieved good results. FskNet [14] proposed 3D-to-2D module and selective kernel selection mechanism, while 3D-SE-DenseNet [15] introduced SE mechanism into 3D-CNN to associate convolutional feature maps of different channels, activate effective information in feature maps, and suppress invalid information. Therefore, using 3D-CNN to extract spatial and spectral information of hyperspectral remote sensing images has become an important direction for hyperspectral remote sensing image classification. We also noticed some methods based on transformers [16-17], which use group spectral embedding and transformer encoder module to model spectral representation [18]. However, these methods have obvious drawbacks, and all these improved transformer methods treat spectral bands or spatial blocks as tokens and encode all tokens, leading to a large amount of redundant computation. However, HSI data already contains a large amount of redundant information, and the accuracy of these methods often cannot compete with that of 3D-CNN, and the computation is also greater than that of 3D-CNN.

3D-CNN has the ability to simultaneously sample in spatial and spectral dimensions. It retains the spatial feature extraction capability of 2D-CNN while ensuring the effectiveness of spectral feature extraction. 3D-CNN can directly process high-dimensional data, eliminating the need for pre-dimensionality reduction of hyperspectral images. However, due to the introduction of the spectral dimension in the convolutional kernel, the parameter count of the feature map is greatly increased, especially when the input spectral dimension is high, resulting in a significant increase in computation and model parameters. In order to reduce the parameter count, decreasing the use of 3D convolution in the network can weaken the feature extraction capability of the convolution kernel, making it difficult to balance between accuracy and computational efficiency, and these models also have high computation complexity, making it difficult to deploy on edge devices with strict latency requirements and limited computing resources. Currently used methods including DFAN, MSDN, 3D-DenseNet, 3D-SE-DenseNet all use dense connections, connecting each layer directly to all previous layers for feature reuse. However, dense connections introduce redundancy in subsequent layers when early features are no longer needed. Therefore, how to introduce better lightweight structures and more effective feature extraction in 3D convolution has become a direction for faster hyperspectral classification.

Using group convolution instead of regular convolution can significantly improve the computational efficiency of deep convolutional networks, which has been widely used in lightweight network design. Standard group convolution divides the input and output channels of the convolutional layer into G mutually exclusive groups and performs regular convolutional operations within a single group, reducing the computational burden by theoretically G times. However, existing group convolution has two major drawbacks: 1. The number of convolution groups in each group and the corresponding convolution kernels are pre-defined and fixed, which hinders the representation ability of group convolution. 2. By introducing sparse connections, the expression ability of convolution is weakened, resulting in reduced performance for difficult samples. Referring to the idea of dynamic learnable networks [22-23], this article introduces learnable group convolution on 3D convolution kernels. The group structure includes the input channel and the group to which the convolution kernel belongs, both of which are learnable, allowing for flexible grouped structures and better representation ability. It is optimized in an end-to-end manner through the overall loss and gradient dynamically determined and updated by backpropagation. Learnable channel numbers and corresponding groups can capture different complementary visual and semantic features of the input image, allowing CNN to learn rich feature representations. Due to the sparsity of hyperspectral data, the sample distribution is uneven in space, and there is a lot of redundant information in the spectral dimension. When extracting high-dimensional and redundant hyperspectral data with 3D convolution, there is also a lot of redundant information between convolution kernels. The LGC module allows 3D-CNN to select channel information with richer semantic features, enabling 3D-CNN to achieve sufficient feature extraction while balancing speed and computational requirements, and is compatible with existing 3D-CNNs, ensuring the integrity of the original network structure. The main contributions of this article are as follows:

1. This article proposes an improved efficient 3D-DenseNet for spatial-spectral hyperspectral image classification based on the LGC module, using 3D-CNN as the base structure and combining dense connections with the LGC module. The ConDenseNet model with LGC achieves good accuracy on the Indian Pines, Pavia University, and KSC datasets with fewer parameters.

2. This article introduces lightweight structural design in 3D-CNN, and LGC improves the shortcomings of existing grouped convolutions. The group structure includes the input channel and the group to which the convolution kernel belongs, both of which are learnable, allowing for flexible grouped structures and better representation ability. It is optimized in an end-to-end manner through the overall loss and dynamically updated gradients determined by backpropagation, greatly reducing the number of parameters, making it applicable to edge devices.

3. Compared with networks that combine various mechanisms, LGCNet is more concise, without complex connections and concatenations, and has less computational complexity. We divided LGCNet into three models: small/base/larger, which remain competitive under lightweight design conditions.

2. Spatial spectral classification method based on lightweight convolutional structure
2.1 Efficient architecture deisgn

As typical sparse-connected convolutions, group convolutions and depthwise separable convolutions are commonly used modules in lightweight structural design. Depthwise separable convolution can be viewed as an extreme case of group convolution, where each convolution kernel corresponds to a channel dimension as shown in Figure 1. AlexNet [24] first used grouped convolution to address memory limitations. ResNeXt [25] further applied grouped convolution and proved its effectiveness. These grouped convolution modules use fixed connections during inference, inevitably weakening the representation ability of convolution. Our proposed LGC effectively enhances feature extraction by using dynamically grouped convolution. Grouped convolution is feature extraction in the channel dimension, and the sparse and redundant spectral characteristics of hyperspectral images are well suited to the pattern of grouped convolution. We did not use depthwise separable convolution because it cannot fully exploit the redundant features of hyperspectral information, as it models only one channel.

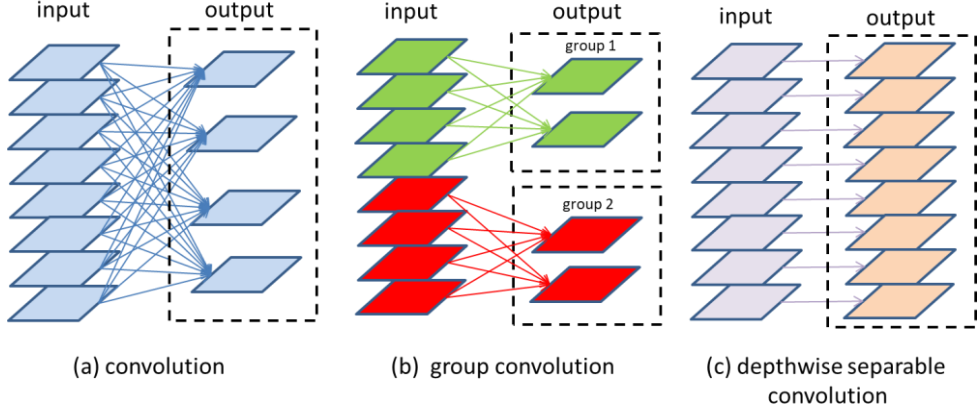

Fig1: Schematic diagram of convolution, group cpnvolution and depthwise separable convolution

Weight pruning [26] or weight quantization [27] can improve the inference efficiency of convolutional networks. Deep neural networks usually have a large number of redundant weights, so they can be pruned or quantized without sacrificing accuracy. Different pruning techniques for convolutional networks may result in different levels of granularity. Fine-grained pruning, such as independent weight pruning [28], can typically achieve high sparsity. LGCNet can generate efficient grouped convolutions that achieve the optimal balance between sparsity and regularity. In the inference stage, LGC designs an effective index reordering strategy, and the reordered index can be obtained offline. LGC can be as efficient as standard group convolution.

2.2 Learnable 3D group convolution.

Hyperspectral image sample data is lacking and exhibits sparse spatial distribution of sample objects, with a large amount of high-dimensional redundant information in the spectral dimension. Although 3D-CNN structures can utilize spatial-spectral information, it is still a challenge to effectively extract spatial-spectral information. Convolutional kernels are the core of convolutional neural networks and are usually considered as an aggregation of spatial and feature dimensions based on local receptive fields. Convolutional neural networks consist of a series of convolution layers, non-linear layers, and downsampling layers, enabling them to capture global receptive fields to describe images. However, learning a high-performance network is difficult, and there have been many efforts to improve network performance from a spatial dimension perspective, such as embedding multiscale information in the Inception structure, aggregating features from different receptive fields to obtain performance gains, and fusing features generated by different Residual blocks for deep feature extraction. However, this type of network has a high training cost, slow convergence speed, and is prone to overfitting on small sample datasets. The convolution kernel of the 3D-CNN also contains a large amount of redundant weights in feature extraction, which contribute little to the final output. This paper introduces a lightweight convolutional structure and dynamic learning mechanism in hyperspectral image processing, and introduces the LGC module in the improved efficient

3D-DenseNet. The LGC module is a structure driven by deep networks that determines how many channels each group should contain and to which group each channel belongs. We use two parameter matrices S and T to parameterize the grouping of input and output channels and automatically optimize these two parameters during training to obtain automatically learned group convolutions. The combination of the LGC module and lightweight group convolutions can greatly reduce model parameter volume and computation complexity. Automatic acquisition of each feature channel's grouping through learning is effective in hyperspectral sparse objects as hyperspectral data often reflects relatively concentrated trends spatially, and combined with the deep feature characteristics of the 3D-DenseNet, it can more effectively extract features.

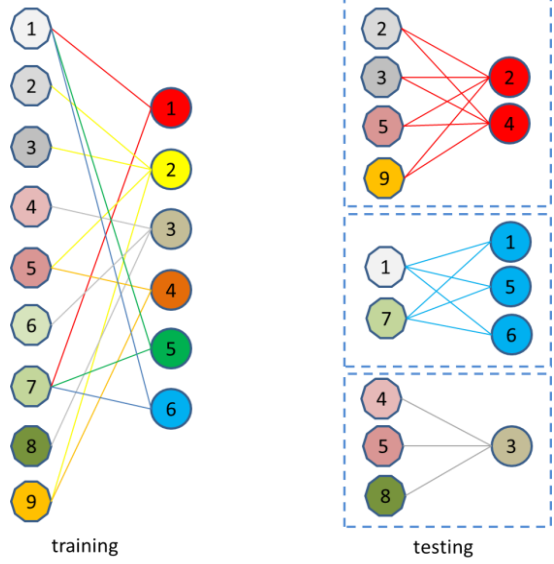

Fig.2 Schematic diagram of learnable convolution training and testing

2.2.1 Training of learnable 3D convolution.

In deep networks, convolution layers are calculated as the convolution between the input feature map and the convolution kernel. Taking the k-th layer as an example, the input can be represented as $X^k = \{X_1^k, X_2^k, \ldots, X_C^k\}$, where C is the number of channels, $x_i^k$ is the i-th feature map, and the convolution kernel of the k-th layer is represented as $W^k = \{w_1^k, w_2^k, \ldots, w_N^k\}$, where N is the number of convolution kernels and $w_i^k$ is a 3D convolution kernel. The output of the convolutional layer is as follows:

$$X^{k+1} = W^k \times X^k = \{w_1^k * X^k, w_2^k * X^k, \ldots, w_N^k * X^k\}$$

× denotes convolution between two sets, while * denotes convolution operation between the convolution kernel and input feature map.

In group convolution, input channels and convolution kernels are divided into G groups, so $X^{k+1}$ can be expressed as:

$$X^{k+1} = \{W_1^k \times X_1^k, W_2^k \times X_2^k, \ldots, W_G^k \times X_G^k\}$$

Usually in standard convolution, input channels and convolution kernels are equally divided into G groups in a hard way, with $\frac{C}{G}$ input channels and $\frac{N}{G}$ convolution kernels in each group, and the speedup ratio is:

$$\frac{Madds(W^k \times X^k)}{Madds(\sum_{i=1}^{G} W_i^k \times X_i^k)} = G$$

This hard-assigned group convolution brings G-fold speedup, but it also has many limitations. Our goal is to design a learnable grouping mechanism that dynamically optimizes the group structure, which is beneficial for speeding up and improving model accuracy. Dynamic grouping is better suited for capturing features that best match the convolution kernel and input channels, and is particularly useful in hyperspectral imaging scenarios.

We represent the grouping structure of the k-th layer as two binary selection matrices for input channels and convolution kernels, denoted as $S^k$ and $T^k$, $S^k$ is a $C \times G$ matrix of channel selections, where each element is defined as:

$$S^k(i, j) = \begin{cases} 1, if \rightarrow x_i^k \in X_j^k, \\ 0, if \rightarrow x_i^k \notin X_j^k, \end{cases} i = [1, C]; j \in [1, G]$$

Here, $S^k(i, j) = 1$ indicates that the i-th input channel is selected into the j-th group. The j-th column of $S^k$ can represent which input channels belong to the j-th group. Therefore,

$$X_j^k = X^k \bullet S^k$$

● represents the major element selection operator.

For the selection of convolution kernels, we define $T^k$ a matrix of selections with dimensions $N \times G$, where each element is defined as:

$$T^k(i, j) = \begin{cases} 1, if \rightarrow w_i^k \in W_j^k, \\ 0, if \rightarrow w_i^k \notin W_j^k, \end{cases} i = [1, N]; j \in [1, G]$$

Here, $T^k(i, j) = 1$ indicates that the j-th convolution kernel is selected into the j-th group. The j-th column of $T^k$ can represent which convolution kernels belong to the j-th group. Therefore,

$$W_j^k = W^k \times T^k$$

The learnable 3D grouped convolution can be represented as

$$\begin{aligned} X^{k+1} &= W^k \times X^k \\ &= \{W_1^k \times X_1^k, W_2^k \times X_2^k, \ldots, W_G^k \times X_G^k\} \\ &= \{W^k \bullet T_1^k \times X^k \bullet S_i^k, \ldots, W^k \bullet T_G^k \times X^k \bullet S_G^k\} \end{aligned}$$

In the above formula, the grouping structure of $W^k, S^k, T^k$ is automatically optimized based on the global objective function. However, the binary function is not differentiable. Therefore, we made a clever approximation: during the training process, we use softmax on S and T along the rows, so that each row contains only one non-zero element, making it differentiable. The LGC grouping structure is automatically optimized rather than specifying hyperparameters manually, based on the loss of the entire network, which means that the grouping structures of all layers in the model can be jointly optimized to extract better features.

2.2.2 Index sorting inference.

After learning the grouping structure, it is usually necessary to reorganize the input channels and convolution kernels for fast inference. One simple method is to add an index layer to reorder the input channels based on the group information, and another index layer to reorder the convolution kernels, as shown in Figure 3(a), and then reorder the output channels back to the original order. However, these frequent memory reordering operations significantly increase the inference time.

Therefore, we propose an effective index reordering strategy as shown in Figure 3(b). First, the convolution kernels are reordered so that the kernels in a group are placed together. Secondly, considering that the input channels are also the output channels of the previous layer, we merge the indices of the previous layer's output and the current layer's input channels into a single index to obtain the correct input channel order, as shown in detail in Figure 3(c). With this design, the memory operations are greatly reduced, and all the reordering indices can be obtained offline, so the efficiency is very high during the inference phase. During inference, LGC can be as efficient as standard grouped convolutions.

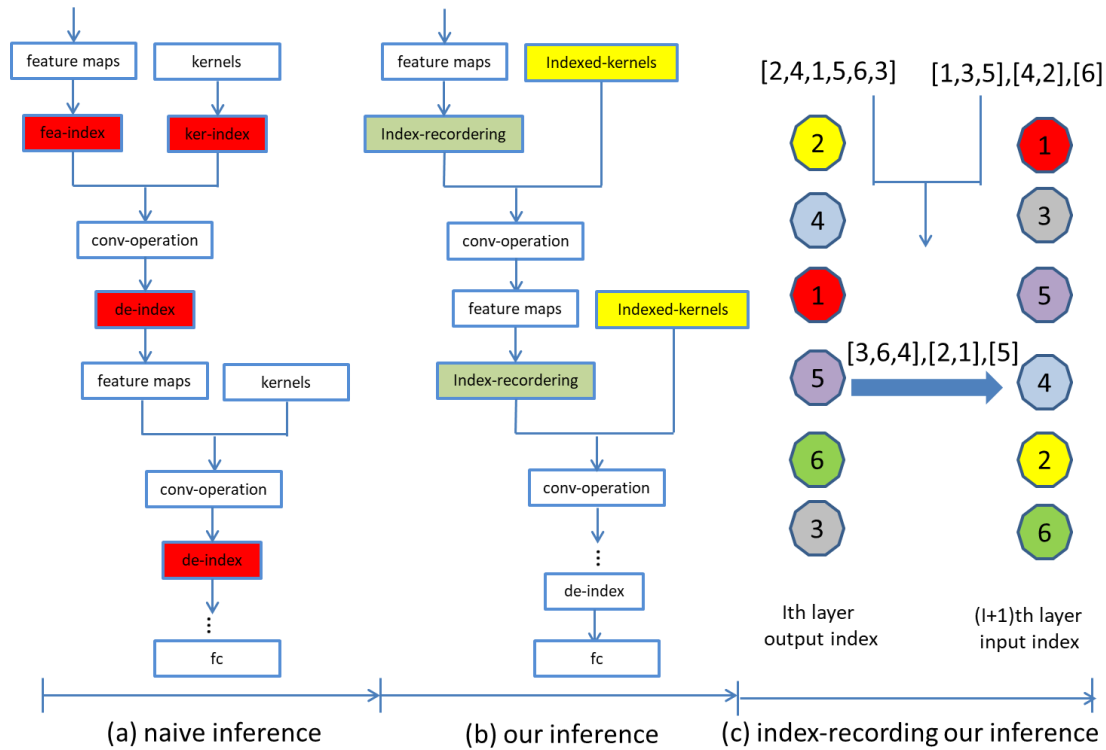

Fig.3 Schematic diagram of index reordering for efficient reasoning. (a) primitive simple reasoning (b) LGC's reasoning (c) index-recording unit

2.3 LGCNet feature extraction framework and model implementation.

We optimized the original 3D-DenseNet in two ways to further simplify the architecture and improve computational efficiency.

2.3.1 Optimization of growth rate.

The original version of DenseNet adds k new feature maps in each layer, where k is a constant known as the growth rate. It has been shown in literature [29] that deep feature extraction in DenseNet tends to rely more on high-level features rather than low-level features and is generally improved by strengthening the short connections. In LGCNet, the growth rate can be gradually increased with depth to increase the proportion of features from deeper layers relative to those from shallower layers. We set the growth rate to   where m is the index of the dense block, and   is a constant. This way of setting the growth rate does not introduce any additional hyperparameters. The increase growth rate strategy places a larger proportion of parameters in the deeper layers of the model, which improves computational efficiency.

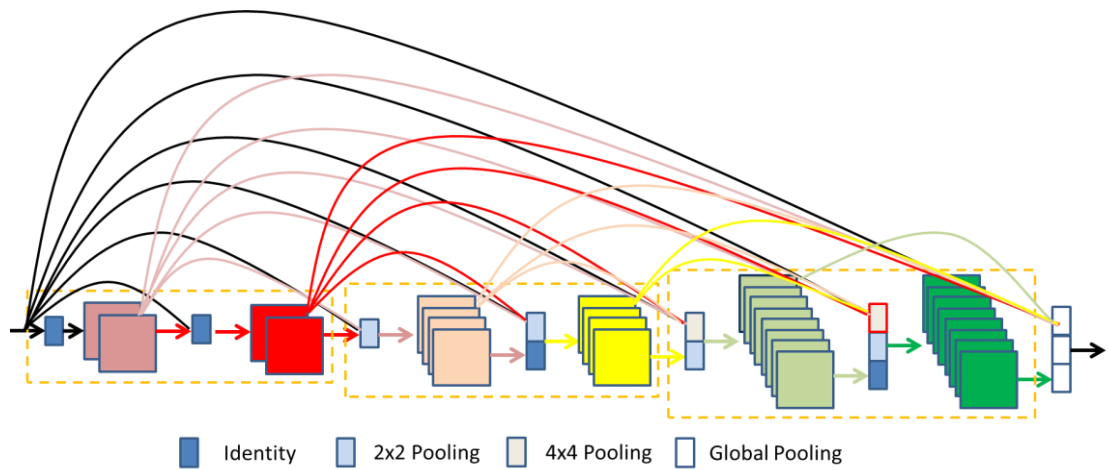

Fig.4 The DenseNet variant in this paper. There are two differences from the original DenseNet: (1) Layers of feature maps of different resolutions are directly connected;(2) The growth rate doubles whenever the feature map size shrinks(More features are generated in the third yellow dense block than in the first block).

2.3.2 Dense block connection optimization.

More features are reused than in the original DenseNet by connecting the input feature map to all subsequent layers in the network, even if those layers are located in different dense blocks (see Figure 4). Since dense blocks have different feature resolutions, average pooling is used to reduce feature map resolution.

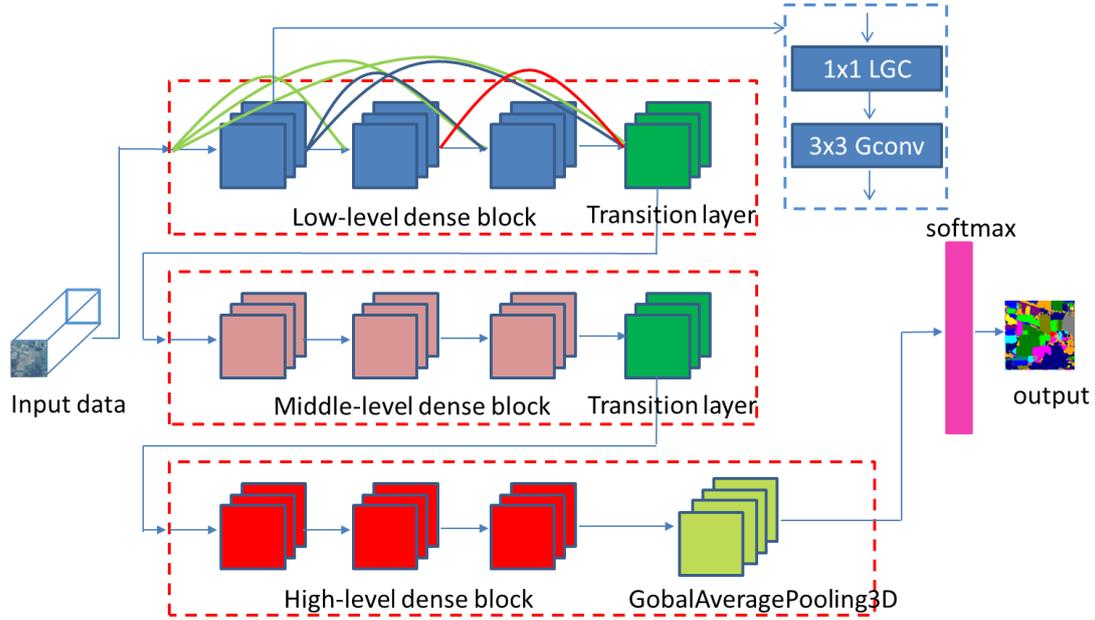

Fig.5 Schematic diagram of the model structure of LGCNet

As shown in Figure 5, the structure diagram of the LGCNet model is presented. We have chosen a smaller model to demonstrate its structure, which includes three stages. The number of dense blocks in each stage is 4, 6, and 8 respectively, with growth rates of 4, 8, and 16.

3. Experiments and analysis.

To evaluate the performance of the LGCNet model, we introduced three of the most representative hyperspectral image datasets: Indian Pines, Pavia University, and KSC. Classification metrics used included overall classification accuracy (OA), average classification accuracy (AA), and Kappa coefficient.

3.1 Dataset.

The Indian Pines dataset was collected in June 1992 from the Indiana Pine Forest experimental area in northwestern Indiana, USA, using the AVIRIS spectral imaging system. The data image has a size of 145x145 pixels, a spatial resolution of 20m, and consists of 220 spectral bands in the wavelength range of 0.4-2.5μm. For our experiments, we removed 20 bands with water vapor absorption and low signal-to-noise ratio, and retained the remaining 200 bands. This dataset includes 16 land cover classes, such as grass, buildings, and crops.

The Pavia University dataset was collected in the Pavia region of northern Italy in 2001 using the ROSIS spectral imaging system. The data image has a size of 610x340 pixels, a spatial resolution of 1.3m, and consists of 115 spectral bands in the wavelength range of 0.43-0.86μm. For our experiments, we removed 12 bands that contained strong noise and water vapor absorption, retaining the remaining 103 bands. This dataset includes 9 land cover classes, such as roads, trees, and rooftops.

The KSC dataset was collected on March 23, 1996, at the Kennedy Space Center in Florida, USA, using the AVIRIS spectral imaging system. AVIRIS collected 224 spectral bands with a bandwidth of 10 nm and a central wavelength range of 400-2500 nm. The spatial resolution of the KSC data, acquired from an altitude of approximately 20 km, was 18m. After removing absorption and low SNR frequency bands, 176 bands were used for analysis, defining 13 classes.

## 3.2 Experimental analysis

LGCNet was trained for 100 epochs on three datasets using the rmsprop optimizer. The training framework saved the model with the highest accuracy on the validation set from 4 training results, with a learning rate of 0.0005. The experiment was conducted on a platform with 8 1080ti GPUs and 11GB of VRAM. For the experiment analysis, LGCNet-small version was used, with a 3-stage structure, 4, 6, and 8 dense blocks in each stage, a growth rate of 4, 8, and 16, and compression ratio of 16.

### 3.2.1 The impact of sample training ratio on classification results

The proportion of the training set is relatively sensitive in small sample size hyperspectral datasets, so the performance of the model under different training, validation, and testing ratios was discussed. LGCNet selected a ratio of 6:1:3 for Indian Pines and Pavia University datasets and a ratio of 4:1:5 for the KSC dataset. We noticed that the model accuracy increased with the increase of the training set, but there was a certain threshold for the increase of the training set, beyond which the accuracy increase was not obvious and may present a oscillating trend. During the experiment, the model chose a neighboring pixel block size of 15.

Table 1 OA, AA and Kappa metrics for different training set ratios on the Indian Pines dataset

| Train ratio | OA | AA | Kappa |
|---|---|---|---|
| 2:1:7 | 96.13 | 94.21 | 95.58 |
| 3:1:6 | 97.61 | 94.10 | 97.27 |
| 4:1:5 | 98.99 | 98.45 | 98.86 |
| 5:1:4 | 99.32 | 98.91 | 99.23 |
| 6:1:3 | 99.45 | 99.28 | 99.38 |

Table 2 OA, AA and Kappa indicators for different training set ratios on the Pavia University dataset

| Train ratio | OA | AA | Kappa |
|---|---|---|---|
| 2:1:7 | 99.58 | 99.38 | 99.44 |
| 3:1:6 | 99.76 | 99.61 | 99.69 |
| 4:1:5 | 99.24 | 99.28 | 99.00 |
| 5:1:4 | 99.87 | 99.86 | 99.83 |
| 6:1:3 | 99.97 | 99.97 | 99.96 |

Table 3 OA, AA and Kappa metrics for different training set proportions on the KSC dataset

| Train ratio | OA | AA | Kappa |
|---|---|---|---|
| 2:1:7 | 98.39 | 97.83 | 98.21 |
| 3:1:6 | 99.59 | 99.38 | 99.55 |
| 4:1:5 | 99.85 | 99.78 | 99.83 |
| 5:1:4 | 99.78 | 99.27 | 99.76 |
| 6:1:3 | 99.84 | 99.80 | 99.82 |

### 3.2.2 The impact of adjacent pixel block size on classification results.

LGCNet pads the input 145x145x200 image to 155x155x200 (taking Indian Pines dataset as an example) and then selects an adjacent pixel block of size MxNxL on the 155x155x200 image. MxN is the spatial sampling size, and L is the spectral dimension. The size of the original input image affects the speed and effectiveness of model training. If the image is too large, it will occupy more VRAM and cause slower training speeds. The size of the adjacent pixel block is an important hyperparameter, but its range should not be too small, otherwise, the receptive field of the convolutional kernel may not be sufficient, and the model may not perform well locally. As shown in Tables 4, 5, and 6, the pixel block size from 7 to 17 improves accuracy on Indian Pines dataset, which is also reflected in Pavia University and KSC datasets. However, as the pixel block range increase, the overall accuracy growth becomes smaller. The training, validation, and testing ratios for Indian Pines, Pavia University, and KSC are 6:1:3, 6:1:3, and 4:1:5, respectively.

Table 4 OA, AA and Kappa indicators under different adjacent pixel block sizes on the Indian pines dataset

| Adjacent pixel block（M=N） | OA | AA | Kappa |
| --- | --- | --- | --- |
| 7 | 92.44 | 92.03 | 91.37 |
| 9 | 94.91 | 95.22 | 94.19 |
| 11 | 98.07 | 98.39 | 97.80 |
| 13 | 98.44 | 98.55 | 98.23 |
| 15 | **99.45** | **99.28** | **99.39** |
| 17 | 99.45 | 99.13 | 99.37 |

Table 5 OA, AA and Kappa indicators under different adjacent pixel block sizes on the Pavia University dataset

| Adjacent pixel block（M=N） | OA | AA | Kappa |
| --- | --- | --- | --- |
| 7 | 99.66 | 99.50 | 99.55 |
| 9 | 99.88 | 99.81 | 99.83 |
| 11 | 99.86 | 99.83 | 99.81 |
| 13 | 99.89 | 99.77 | 99.85 |
| 15 | **99.97** | **99.97** | **99.96** |
| 17 | 99.95 | 99.95 | 99.93 |

Table 6 OA, AA and Kappa indicators under different adjacent pixel block sizes under the KSC dataset

| Adjacent pixel block（M=N） | OA | AA | Kappa |
| --- | --- | --- | --- |
| 7 | 94.42 | 93.44 | 93.79 |
| 9 | 97.77 | 96.72 | 97.52 |
| 11 | 98.15 | 97.72 | 97.94 |
| 13 | 98.81 | 98.81 | 98.96 |
| 15 | **99.85** | **99.78** | **99.83** |
| 17 | 99.84 | 99.80 | 99.82 |

3.2.3 Network parameter

We divided LGCNet into three types: small, base, and larger. The following table shows the testing results on Indian Pines dataset, where OA, AA, and Kappa are evaluation metrics. The larger model has the highest accuracy, although it also has the most parameters.

Table 7 LGCNet network model parameter configuration and OA, AA and Kappa indicators

|  | Stages/dense block | Rate. | OA | AA | Kappa |
| --- | --- | --- | --- | --- | --- |
| Lgcnet-small | 4,6,8 | 4,8,16 | 99.45 | 99.28 | 99.39 |
| Lgcnet-base | 6,8,10 | 8,16,32 | 99.58 | 99.53 | 99.83 |
| Lgcnet-larger | 10,10,10 | 8,16,32 | 99.88 | 99.53 | 99.86 |

Using the thop library, we obtained the number of parameters (params) and computation of FLOPs (gflops) on the Indian Pines dataset, where the spectral dimension is 200.

Table 8 Comparison of parameters and calculations of different models on the Indian Pines dataset

| | 3D-CNN | 3D-DenseNet | HybridSN | Lgcnet-small | Lgcnet-base | Lgcnet-larger |
|---|---|---|---|---|---|---|
| Params | 16394652 | 2562452 | 5503108 | 156856 | 882272 | 1834848 |
| Flops | 81959692 | 5234500 | 11005179 | 6898600 | 36199104 | 72421584 |

3.3 Experimental results and analysis.

On the Indian Pines dataset, the input size of LGCNet is 15x15x200. On the Pavia University dataset, the input size of LGCNet is 15x15x103. On the KSC dataset, the input size of LGCNet is 15x15x176. We evaluated Lgcnet-small/base/larger and compared them with SAE, SSRN, 3D-CNN, 3D-SE-DenseNet, and Spectralformer, as shown in Tables 9, 10, and 11. Among them, all three combinations of LGCNet showed overall leading accuracy.

Table 9 Comparison of the classification accuracies (%) of different methods for the Indian Pines dataset

| No | SAE | SSRN | 3D-CNN | 3D-SE-DenseNet | Spectralformer | Lgcnet-small | Lgcnet-base | Lgcnet-larger |
|---|---|---|---|---|---|---|---|---|
| 1 | 81.82 | 100 | 96.88 | 95.87 | 70.52 | 100 | 100 | 100 |
| 2 | 82.16 | 99.85 | 98.02 | 98.82 | 81.89 | 98.85 | 99.92 | 99.92 |
| 3 | 77.54 | 99.83 | 97.74 | 99.12 | 91.30 | 99.22 | 99.87 | 99.87 |
| 4 | 68.11 | 100 | 96.89 | 94.83 | 95.53 | 98.58 | 100 | 99.09 |
| 5 | 94.36 | 99.78 | 99.12 | 99.86 | 85.51 | 99.48 | 100 | 100 |
| 6 | 94.45 | 99.81 | 99.41 | 99.33 | 99.32 | 98.89 | 99.56 | 100 |
| 7 | 94.70 | 100 | 88.89 | 97.37 | 81.81 | 96.88 | 95.83 | 100 |
| 8 | 94.36 | 100 | 100 | 100 | 75.48 | 99.81 | 100 | 100 |
| 9 | 82.56 | 0 | 100 | 100 | 73.76 | 100 | 100 | 94.44 |
| 10 | 81.28 | 100 | 100 | 99.48 | 98.77 | 99.91 | 99.78 | 99.67 |
| 11 | 84.47 | 99.62 | 99.33 | 98.95 | 93.17 | 99.46 | 99.82 | 100 |
| 12 | 83.77 | 99.17 | 97.67 | 95.75 | 78.48 | 99.18 | 100 | 99.46 |
| 13 | 96.42 | 100 | 99.64 | 99.28 | 100 | 100 | 100 | 100 |
| 14 | 92.27 | 98.87 | 99.65 | 99.55 | 79.49 | 99.81 | 100 | 100 |
| 15 | 80.63 | 100 | 96.34 | 98.70 | 100 | 100 | 100 | 100 |
| 16 | 81.82 | 98.51 | 97.92 | 96.51 | 100 | 97.38 | 97.73 | 100 |
| OA | 85.47±0.58 | 99.62±0.00 | 98.23±0.12 | 98.84±0.18 | 81.76 | 99.45±0.30 | 99.85±0.04 | 99.88±0.06 |
| AA | 86.31±1.14 | 93.46±0.50 | 98.80±0.11 | 98.42±0.56 | 87.81 | 99.28±0.33 | 99.53±0.23 | 99.53±0.50 |
| K | 83.42±0.66 | 99.57±0.00 | 97.96±0.53 | 98.60±0.16 | 79.19 | 99.38±0.34 | 99.83±0.05 | 99.86±0.06 |

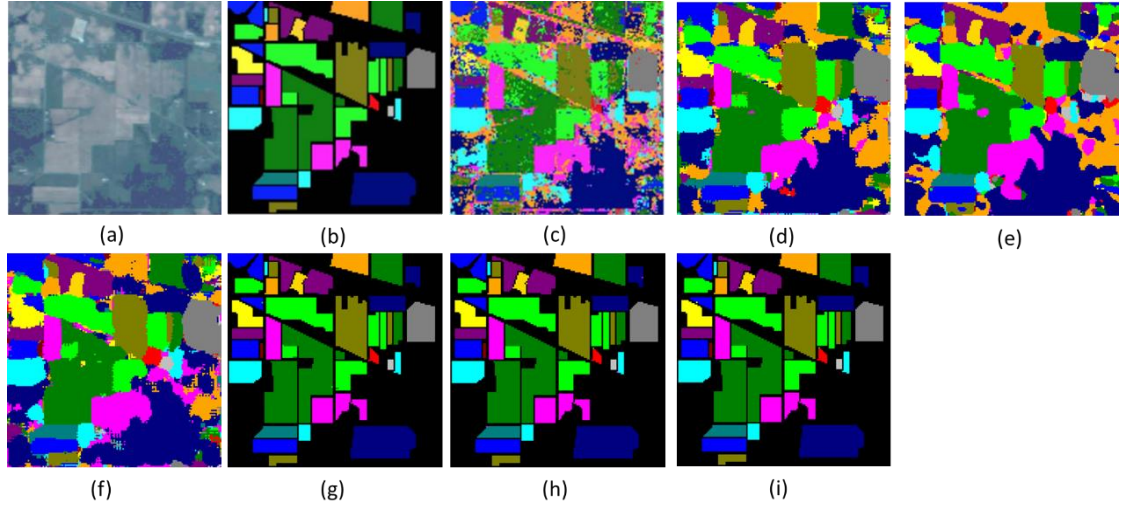

Fig.14. Classification results of the best models for the Indian Pines dataset:(a)false color image, (b)ground-truth labels, (c)-(i)classification results of the SAE, SSRN, 3D-CNN, 3D-SE-DenseNet-BC,Lgcnet-small,Lgcnet-base,Lgcnet-larger.

Table 10 Classification accuracy(%) of different methods for the Pavia University dataset

| No | SAE | SSRN | 3D-CNN | 3D-SE-DenseNet | Spectralformer | Lgcnet-small | Lgcnet-base | Lgcnet-larger |
|----|-----|------|--------|----------------|----------------|--------------|-------------|---------------|
| 1  | 98.56 | 89.93 | 99.96 | 99.32 | 82.73 | 99.95 | 100 | 100 |
| 2  | 99.77 | 86.48 | 99.99 | 99.87 | 94.03 | 100 | 100 | 100 |
| 3  | 99.05 | 99.95 | 99.64 | 96.76 | 73.66 | 100 | 99.88 | 100 |
| 4  | 99.98 | 95.78 | 99.83 | 99.23 | 93.75 | 100 | 100 | 100 |
| 5  | 99.90 | 97.69 | 99.81 | 99.64 | 99.28 | 99.94 | 100 | 100 |
| 6  | 99.03 | 95.44 | 99.98 | 99.80 | 90.75 | 100 | 100 | 100 |
| 7  | 99.71 | 84.40 | 97.97 | 99.47 | 87.56 | 100 | 100 | 100 |
| 8  | 97.53 | 100 | 99.56 | 99.32 | 95.81 | 99.82 | 100 | 100 |
| 9  | 99.86 | 90.58±0.18 | 100 | 100 | 94.21 | 100 | 100 | 100 |
| OA | 98.47±0.41 | 92.99±0.39 | 99.79±0.01 | 99.48±0.02 | 91.07 | 99.97±0.02 | 99.99±0.00 | 100±0.00 |
| AA | 99.28±0.31 | 87.21±0.25 | 99.75±0.15 | 99.16±0.37 | 90.20 | 99.97±0.03 | 99.99±0.01 | 100±0.00 |
| K  | 98.97±0.21 | 87.24 | 99.87±0.27 | 99.31±0.03 | 88.05 | 99.96±0.03 | 99.99±0.00 | 100±0.00 |

Table 11 Comparison of the classification accuracies (%) of different methods for the Kennedy Space Center dataset

| No | Lgcnet-small | Lgcnet-base | Lgcnet-larger |
|----|--------------|-------------|---------------|
| 1  | 99.94 | 100 | 100 |
| 2  | 98.82 | 99.74 | 99.18 |
| 3  | 99.41 | 100 | 99.74 |
| 4  | 99.80 | 100 | 100 |
| 5  | 99.71 | 99.19 | 98.06 |
| 6 | 99.79 | 100 | 100 |
| 7 | 100 | 100 | 100 |
| 8 | 99.88 | 100 | 99.08 |
| 9 | 99.90 | 99.86 | 100 |
| 10 | 100 | 100 | 99.51 |
| 11 | 100 | 100 | 100 |
| 12 | 99.90 | 100 | 100 |
| 13 | 100 | 100 | 100 |
| OA | 99.85±0.10 | 99.95±0.07 | 99.77±0.05 |
| AA | 99.78±0.16 | 99.91±0.13 | 99.66±0.04 |
| K | 99.83±0.12 | 99.83±0.08 | 99.74±0.06 |

4.Conclusion

In this paper, we introduced learnable 3D grouped convolution with a group structure that includes both input channels and convolutional kernels. LGC allows for flexible grouping structures and produces better representation capabilities. It is optimized end-to-end through the overall loss and dynamically updated gradients by backpropagation. Learnable channel numbers and corresponding groups can capture different complementary visual and semantic features of input images, enabling the CNN to learn rich feature representations. The sparsity and uneven sample distribution in space, as well as redundancy in the spectral dimension, exist in high-dimensional hyperspectral data. When extracting high-dimensional and redundant hyperspectral data, 3D convolutions contain a large amount of redundant information between convolutional kernels. The LGC module enables 3D-CNN to select channel information with richer semantic features, discard non-activated regions, complete sufficient feature extraction, and balance the demands for speed and computing power while retaining the original structure of the network. Good results have been achieved on the Indian Pines, Pavia University, and KSC datasets.